\documentclass[11pt]{article}
\usepackage{acl2017}
\usepackage{times}
\usepackage{url}
\usepackage{latexsym}
\usepackage{xspace}
\usepackage{booktabs}
\usepackage{color}
\usepackage{graphicx}
\usepackage{tabulary}
\usepackage{enumitem}
\usepackage{amsfonts}

  {\begin{itemize}[topsep=0pt, partopsep=0pt] %
    \setlength{\itemsep}{0pt}%
    \setlength{\parskip}{0pt}%
    }%
  {\end{itemize}}
  \usepackage{subcaption}
\usepackage{lipsum}

  \usepackage{color}
  {\begin{enumerate}≈ \footnotesize%
    \setlength{\itemsep}{0pt}%
    \setlength{\parskip}{0pt}%
    }%
  {\end{enumerate}}

\usepackage{microtype}
\usepackage{multirow}
\usepackage{verbatim}
\usepackage{amsmath,amsthm,amssymb}
\usepackage{array}
\usepackage[scaled=0.86]{helvet}
\usepackage{ifthen}
\usepackage{courier}
\usepackage[linesnumbered,vlined,ruled]{algorithm2e}

\newcommand{\captionfonts}{\small}
\makeatletter  % Allow the use of @ in command names
\long\def\@makecaption#1#2{%
  \vskip\abovecaptionskip
  \sbox\@tempboxa{{\captionfonts #1: #2}}%
  \ifdim \wd\@tempboxa >\hsize
    {\captionfonts #1: #2\par}
  \else
    \hbox to\hsize{\hfil\box\@tempboxa\hfil}%
  \fi
  \vskip\belowcaptionskip}
\makeatother   % Cancel the effect of \makeatletter

\belowdisplayskip \abovedisplayskip

\DeclareMathOperator*{\argmax}{arg\,max}

\aclfinalcopy

\newcommand{\eos}{{\it EOS}\xspace}
\newcommand{\sts}{{{\textsc{Seq2Seq}}}\xspace}

\title{Learning to Decode for Future Success}
\author{
Jiwei Li, Will Monroe and Dan Jurafsky\\
Computer Science Department, Stanford University, Stanford, CA, USA \\
{\tt jiweil,wmonroe4,jurafsky@stanford.edu} 
}
\date{}

\begin{document}
\maketitle

\begin{abstract}
We introduce a simple, general strategy to manipulate the behavior of a neural decoder that enables it to generate outputs that have specific properties of interest (e.g., sequences of a pre-specified length).
The model can be thought of as a simple version of the actor-critic model that uses an interpolation of the actor (the MLE-based token generation policy) and the critic (a value function that estimates the future values of the desired property) for decision making. 
We demonstrate that 
the approach is 
able to incorporate a variety of properties that cannot be handled by  standard neural sequence decoders, such as sequence length and backward probability (probability of sources given targets), in addition to yielding consistent improvements in abstractive summarization and machine translation when the property to be optimized is BLEU or ROUGE scores. 
\end{abstract}
\section{Introduction}

Neural generation models \cite{sutskever2014sequence,bahdanau2014neural,cho2014learning,kalchbrenner2013recurrent}
learn to map source to target sequences  in applications such as machine translation 
\cite{sennrich2015neural,gulcehre2015using}, conversational response generation 
\cite{vinyals2015neural,sordoni2015neural}, 
abstractive summarization \cite{nallapatiabstractive,rush2015neural}.

Neural generation models are standardly trained by maximizing the likelihood of target sequences  given source sequences in a training dataset.
At test time, a decoder incrementally generates a sequence with the highest probability  using search strategies such as beam search. 
This locally incremental nature of the decoding model  leads to 
the following issueL
 Decoders cannot be tailored to generate target sequences with specific properties of interest, such 
as pre-specified length constraints \cite{shao15,shi2016neural}, which might be useful in tasks
like conversational response generation or non-factoid question answering,
and cannot deal with important objectives, such as the mutual information between sources and targets \cite{li2016diversity},
that require knowing the full target sequence in advance.

%\footnote{ One simple workaround is to split the training set by length and train separate models. However, this results in each subset having only a small amount of training data, leading to lower-quality output from all the models.} 

%Sometimes this is because some of these features are not explicitly modeled at training time; for example, 
%current decoding algorithms are
%incapable of 
 %generating sequences that obey pre-specified length constraints 
 %\cite{shao15,shi2016neural}.\footnote{Length might not be an issue in machine translation or image caption generation, but is an aspect of interest in conversational response generation and non-factual question answering. One simple workaround is to split the training set by length and train separate models. However, this results in each subset having only a small amount of training data, leading to lower-quality output from all the models.} 
%Other times it is because the corresponding objectives cannot be handled by search algorithms---e.g., 
%mutual information between sources and targets \cite{li2016diversity}. 

To address this issue, we propose a general strategy 
that allows the decoder to incrementally generate  output sequences that, when complete,
will have specific properties of interest.
Such properties can take various forms, such as length, diversity, mutual information between sources and targets, and  BLEU/ROUGE scores. 
The proposed framework integrates two models: the standard seq2seq model, trained to incrementally predict the next token,
and a future output estimation model, trained to estimate future properties solely from a prefix string (or the representation associated with this string),
and incorporated into the decoder to encourage it to make decisions that lead to better long-term future outcomes. 

Making decoding decisions based on future success resembles the central idea of reinforcement learning (RL), that of training a policy that leads to better long-term reward. Our work is thus related to a variety of recent work  inspired by or using reinforcement learning (e.g., REINFORCE or actor-critic models) for sequence generation \cite{wiseman2016sequence,shen2015minimum,bahdanau2016actor,marc2016sequence}. 
The proposed model can be viewed as a simpler but more effective version of the actor-critic RL model \cite{konda1999actor} in sequence generation: it does not rely on the critic to update the policy (the actor), but rather, uses a linear interpolation of the actor (the policy) and the critic (the value function) to make final decisions. 
 Such a strategy
 comes with the following benefits:
 (1) It 
  naturally avoids the known problems such as  large variance and 
 instability
  with the use of reinforcement learning in tasks with enormous search spaces like sequence generation.
  As will be shown in the experiment sections, 
  the simplified take on reinforcement without policy updates
yields consistent improvements, not only outperforming standard \sts models, but also the RL models themselves in a wide range of sequence generation tasks; 
(2)
  training RL-based generation models using specific features like sequence length as rewards not only increases the model's instability but may also 
lead to suboptimal generated utterances, for example, sequences that satisfy a length constraint but are irrelevant, incoherent or even ungrammatical.\footnote{A workaround is to use the linear interpolation of the MLE-based policy and the value function for a specific property as a reward for RL training. This strategy comes with the following disadvantages: it
requires training different models for different interpolation weights, and again, suffers from large training variance.}

We study how to incorporate different properties into the decoder  different properties of the future output sequence: (1)
sequence length: the approach provides the flexibility of controlling the output length, 
which in turns addresses sequence models' bias towards generating short sequences \cite{sountsov2016length};
(2) mutual information between sources and targets: the approach enables modeling the bidirectional dependency between sources and targets at each decoding time-step, significantly improving response quality on a task of conversational response generation and (3) 
the properties can also take the form of the BLEU and ROUGE scores, 
yielding consistent improvements in machine translation and summarization, 
yielding the state-of-the-art result on the IWSLT German-English translation task.

 \begin{comment}
 \footnote{\newcite{wiseman2016sequence} report higher BLEU scores achieved by MLE based sequence training than RL based approaches.}
 \section{Related Work} \label{sec:related}
\paragraph{Reinforcement Learning}
\paragraph{Decoding Algorithms}
This paper is related to recent attempts to improve neural decoders. 
\newcite{cho2016noisy}
proposed 
 a meta-algorithm that
runs in parallel many chains of the noisy version of an inner decoding algorithm.
\newcite{li2016simple} and 
\newcite{vijayakumar2016diverse} 
 proposed decoding algorithms that promote diversity
 by penalizing hypotheses that are
siblings in the beam search and
 by using a diversity-augmented objective, respectively.
\newcite{shao15}  
used a stochastic search algorithm 
that reranks the hypothesis segment by segment, which injects diversity earlier in the decoding process.
 \end{comment}

\section{Model Overview} \label{sec:overview}
In this section, we first review the basics of 
training and decoding in 
standard neural generation models. Then we give a sketch of the proposed model. 
\subsection{Basics}
Neural sequence-to-sequence (\sts) generation models aim to generate a sequence of tokens $Y$ given input sequence $X$. 
%The input $X$ can be a token sequence or an image, among other possibilities. 
Using recurrent nets, LSTMs \cite{hochreiter1997long} or CNNs \cite{krizhevsky2012imagenet,kim2014convolutional}, $X$ is first mapped to a vector representation,  which is then used as the initial input to the decoder.
A neural generation model 
 defines a distribution over outputs by sequentially predicting tokens using a softmax function:
\begin{equation*}
\begin{aligned}
p(Y|X)
&=\prod_{t=1}^{n_Y}p(y_t|X,y_{1:t-1})\\
%% MG: \in [1,N] is for real numbers
%&=\prod_{t\in [1,n_y]}p(y_t|x_1,x_2,...,x_t,y_1,y_2,...,y_{t-1})\\
%&=\prod_{t\in [1,n_y]}\frac{\exp(f(h_{t-1},e_{y_t}))}{\sum_{y'}\exp(f(h_{t-1},e_{y'}))}
\end{aligned}
\label{equ-lstm}
\end{equation*}
Decoding typically seeks to find the maximum-probability sequence $Y^*$ given input $X$:
\begin{equation}
Y^*=\argmax_{Y}p(Y|X)
\end{equation}
The softmax function that computes
$p(y_t|X,y_{1:t-1})$ takes as input the hidden representation at time step $t-1$, denoted by $h_{t-1}$. The hidden representation $h_{t-1}$ is computed using a recurrent net that combines the previously built representation $h_{t-2}$ and the word representation $e_{t-1}$ for word $y_{t-1}$.  
It is infeasible to enumerate the large space of possible sequence outputs, so beam search is normally employed to find an approximately optimal solution. 
Given a partially generated sequence $y_{1:t-1}$, the score for choosing token $y_t$ (denoted by $S(y_t)$) is thus given by
\begin{equation}
S(y_t)=\log p(y_t|h_{t-1})
\label{logp}
\end{equation}
\subsection{The Value Function   Q}
The core of the proposed  architecture is to train a future outcome prediction function (or value function) $Q$,
which estimates the future outcome of taking an action (choosing a token) $y_t$ in the present.
The  function $Q$
 is then incorporated into  $S(y_t)$ at each decoding step to push the model to generate outputs that lead to future success. This yields the following definition for the score  $S(y_t)$ of taking action $y_t$:
\begin{equation}
S(y_t)=\log p(y_t|h_{t-1})+\gamma Q(X,y_{1:t})
\end{equation}
where $\gamma$ denotes the hyperparameter controlling the trade-off between the local probability prediction $p(y_t|h_{t-1})$ and the 
 value function $Q(X,y_{1:t})$.
The input to $Q$ can take various forms, such as the vector representation of the decoding step after $y_t$ has been considered (i.e., $h_t$) or the raw strings ($X$ and $y_{1:t}$).\footnote{One can think of $h_t$ as the output of a function that takes as input $X$ and $y_{1:t}$.} 

$Q$ can be trained either jointly with or independently of the \sts model. 
When training $Q$, we provide it with source-target pairs $(X,Y)$, where $Y$ is a full sequence. $Y=\{y_1,y_2,...,y_N\}$ can either be sampled or decoded
 using a trained model (making $Q$ dependent on the pre-trained \sts model) or can be taken from the training set (making $Q$ independent of the \sts model). However, $Y$ must always be a full sequence. 
The future outcome of generating each of the tokens of $Y$ ($y_1$, $y_2$, ..., $y_N$) is the feature score (BLEU, length, mutual information, etc.) associated with the full sequence $Y$, denoted by $q(Y)$. The future outcome function $Q$ is trained to predict $q(Y)$ from $(X,y_{1:t})$, where $1\leq t\leq N$.

$Q$ estimates the long-term outcome of taking an action $y_t$. It is thus similar to 
the value function in Q-learning, the role of the critic in actor-critic reinforcement learning \cite{sutton1988learning,grondman2012survey}, the value network
for position evaluation in the Monte-Carlo tree search
of AlphaGo \cite{silver2016mastering}, or the 
h* function in $A^*$ search. \cite{och2001efficient}.

In this paper, {\it value function} and {\it future outcome prediction function} and $Q$  are interchangeable

In the sections below, we will describe how to adapt this general framework to various features with different properties and different kinds of input to the future outcome prediction function.

\section{Q for Controlling Sequence Length} \label{sec:length}
For tasks like machine translation, abstractive summarization and image caption generation, the information required to generate the target sequences is already embedded in the input. Usually we don't have to worry about the length of targets, since the model can figure it out itself; this is a known, desirable property of neural generation models \cite{shi2016neural}. 
\begin{figure*}[ht]
\centering
\includegraphics[width=5.5in]{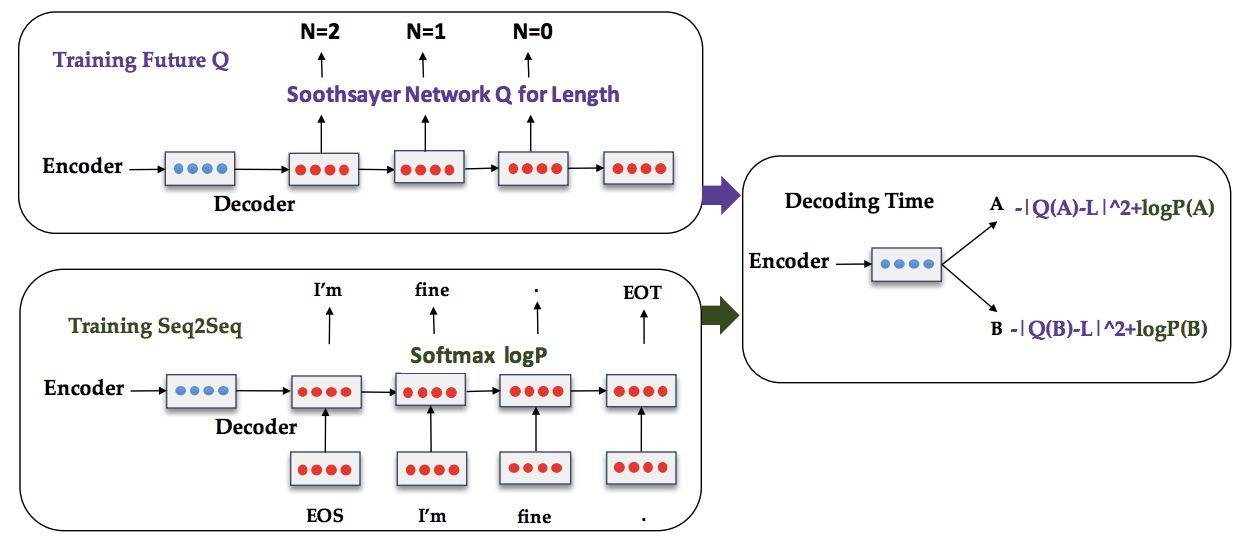}
\caption{An illustration of the proposed future length prediction model. ${\bf N}$ denotes the number of words left to generate and $L$ denotes the pre-specified sequence length.}
\end{figure*}

However, for tasks like conversational response generation and non-factoid question answering, in which there is no single correct answer, 
it is useful to be able to control the length of the targets. 
Additionally, in tasks like conversational response generation,
\sts models have a strong bias towards generating short sequences \cite{sountsov2016length}.
This is because the standard search algorithm at decoding time can only afford to explore very a small action space.
As decoding proceeds, only a small number of hypotheses can be maintained. 
By Zipf's law, short sequences are significantly more frequent than longer ones. Therefore, the prefixes of shorter responses are usually assigned higher probability by the  model. This makes prefixes of longer sequences fall off the beam after a few decoding steps, leaving only
 short sequences. 

One can still force the model to keep generating tokens (simply by prohibiting the \eos token and forcing the decoding to proceed). However, since the previous decoded tokens were chosen with a shorter sentence in mind, artificially lengthening the response this way will result in low-quality responses. In particular, problems arise with repetition (``{\it no, no, no, no, no, no}") or incoherence ({\it ``i like fish, but i don't like fish but I do like fish"}). 
\subsection{Training Q for Sequence Length}
\newcite{shao15} give 
one efficient method of generating long sequences, consisting of 
a stochastic search algorithm 
and segment-by-segment reranking of hypotheses. 
The fundamental idea is to keep a
diverse list of hypotheses on the beam 
 and remove those that are similar 
 to each other,
 so as to explore the space more adequately.
While more adequately 
exploring the search space can increase the likelihood of generating long sequences,
since the beam is more likely to include a prefix of a long sequence, this method doesn't offer direct control over sequence length. 
\begin{comment}
We propose two different models to handle the length control issue, one interfering with the decoding process by linearly adding an additionally trained length-prediction reranker, the other directly interfering with the training processes. 
{Training  a Length-Predictor}
\end{comment}
Length information  seems to be embedded in the hidden representations of neural models in some implicit way \cite{shi2016neural}.
We therefore build another neural model  to expose this length information and use it to estimate the number of words left to be generated.  
\begin{table*}
\small
\centering
\begin{tabular}{ll}\hline
{\bf Input}&it s an 8 symbol address .\\
{\bf SBS}&i don 't know where it came from but it s not the same as it used to be .\\
{\bf FP($\lambda$=1)}&you know i think i ve seen it once and i ve seen it in the past too ! .\\
{\bf FP($\lambda$=5)}&but we don t know who she was at the scene but that s a good idea ! ? !\\\hline
{\bf Input}&come on pollyanna let 's go .\\
{\bf SBS}&come on let 's go let 's go go go go go go go go go go go go go\\
{\bf FP($\lambda$=1)}&you want me to take you to your house and come back and get me some ice cream ? !\\
{\bf FP($\lambda$=5)}&you want me to take the stairs and go with you ... and get the hell out of here ! !\\\hline
{\bf Input}&that 's hardly a winning attitude .\\
{\bf SBS}&the only way that we could find out is that we 're in the middle of the game ! ! !\\
{\bf FP($\lambda$=1)}&but i think the most important thing is that we 're both in the same class together . . .\\
{\bf FP($\lambda$=5)}&but the other thing is i think you can use it as an excuse for your own personal gain .\\\hline
{\bf Input}&i see and you know why ?\\
{\bf SBS}&because i don 't know who i am or where i am or where i am or where i am\\
{\bf FP($\lambda$=1)}&why would the people think i 'm the only guy who could get a job in the middle east ?\\
{\bf FP($\lambda$=5)}&i know the guy i saw at the scene ... and he said it was my father 's birthday !\\\hline
{\bf Input}&but venus was no easy target .\\
{\bf SBS}&it was the only way to make it to the top of the island . . . . . . but it wasn 't\\
{\bf FP($\lambda$=1)}&i think we have the right to be in the middle of some sort of a trap . . .\\
{\bf FP($\lambda$=5)}&the only reason i left here to save you ... was because i didn 't care who they are !\\\hline
{\bf Input}&i 'm not afraid of her .\\
{\bf SBS}&i 'm afraid she 's afraid of the dark ... ... but i 'm afraid she 's afraid of me .\\
{\bf FP($\lambda$=1)}&you don 't like to tell people you 're just a child and you don 't know her ? !\\
{\bf FP($\lambda$=5)}&i 'll be in a very awkward moment of her time and i 'm afraid she 'll hurt us again\\\hline
\end{tabular}
\caption{Sample of responses generated by standard beam search (denoted by {\it SBS}) and the  future length prediction ({\it FP}) algorithm with two different values of $\lambda$. We force each decoding algorithm to generate responses with length 20. More examples are shown in Table~\ref{length20} (Appendix).}
\label{sample-length}
\end{table*}

Given a pre-trained sequence-to-sequence model, an input sequence $X$, and a target $Y=\{y_{1},y_2,...,y_N\}$, where $N$ denotes the length of $y$, we first run a forward pass to compute the hidden representation $h_t$ associated with each time step on the target side  ($1\leq t\leq N$). 
Then we build a regression model $Q(h_t)$, which takes as input $h_t$ to predict the length of the remaining sequence, i.e., $N-t$. The model first passes $h_t$ to two non-linear layers, on top of which is a linear regression model which outputs the predicted number of tokens left to decode. 
The regression model is
optimized by 
minimizing the mean squared loss between the predicted sequence length and
the gold-standard length
 $N-t$ on source--target pairs taken from the training set. 
 
 \subsection{Decoding}
Given an input $X$, suppose that we wish to generate a target sequence of a pre-specified length $N$. 
At decoding time step $t-1$, 
we first obtain the vector representation $h_{t-1}$ for the current time step. The score used to rank choices for the next token $y_{t}$ is a linear combination of the log probability outputted from the sequence model and the mean square loss between the number of words left to generate ($N-t$) and the output from $Q(h_t)$:
\vspace*{-10pt}
\begin{equation}
\begin{aligned}
&y_t=\argmax_y \log p(y_{1:t}|X)\\[-4pt]
&~~~~~~~~~~~~~~~~~~~~~~~-\lambda||(N-t)-Q(h_t)||^2
\end{aligned}
\label{eq3}
\end{equation}
where $y_{1:t}$ is the concatenation of $y_t$ and previously decoded (given) sequence $y_{1:t-1}$, and $h_t$ is obtained from the sequence model by combining $h_{t-1}$ and the word representation of $y_t$ as if it were the true next token. 
$\lambda$ is a hyperparameter controlling the influence of the future length estimator. 
\subsection{Experiments}
\label{length-e}
We evaluate the proposed model on the task of open-domain conversation response generation \cite{vinyals2015neural,sordoni2015neural,serban2015building,mei2016coherent,serban2015hierarchical}, in which 
a model must predict the next turn of a dialogue given the preceding
ones. 
We use 
the OpenSubtitles 
(OSDb) dataset \cite{tiedemann2009news}.
\begin{comment}
\footnote{ 
The
dataset does not specify which character speaks
each subtitle line, which prevents us from inferring
speaker turns. Following Vinyals et al. (2015), we
make the simplifying assumptions that each line of
subtitle constitutes a full speaker turn and that consecutive
turns belong to the same conversation. This introduces
a degree of noise, since in reality consecutive lines
may or may not be spoken by the same character,
and might not appear in the same conversation or even scene.}
We train attention models using the same structure as in the previous sections. 
The trained future length prediction model obtains 
a 
mean squared loss of 11.7 on a held-out dev set; for comparison, the loss is 38.3 for a model that always predicts the remaining length to be the half of the average length of targets from the training dataset. 
\end{comment}

We compare the proposed model with the standard \sts beam search (SBS) decoder. 
We first group test dialogue pairs by target length and decode each group. 
At decoding time, for an input with a gold target of length $L$,
we force the model to generate an output of length $L$. This can be achieved by selecting a hypothesis that predicts an \eos token at time step $L+1$.\footnote{If multiple hypotheses satisfy this requirement, we pick the one with the largest likelihood.} If no \eos is predicted at time step $L+1$, we continue decoding and stop once an \eos token is generated.  
We report BLEU scores on the concatenation of the outputs from each length cluster.\footnote{In this setup, both algorithms are allowed to know the length of the gold-standard targets.
The results from different models are thus comparable. 
 This is to remove the effect of target length on the evaluation metrics (all metrics employed are sensitive to target length).}

We also report adversarial success ({\it AdverSuc}) and {\it machine-vs-random} accuracy, 
evaluation metrics proposed in \newcite{add}. 
   Adversarial success refers to the percentage of machine-generated responses that are able to fool a trained evaluator model into believing that they are generated by a human; {\it machine-vs-random} accuracy denotes the accuracy of a (different) trained evaluator model at distinguishing between machine-generated responses and randomly-sampled responses.\footnote{The estimators for {\it AdverSuc} and {\it machine-vs-random} accuracy are trained using a hierarchical network \cite{serban2016building} See \newcite{add} for details.} 
   \begin{comment}
    one LSTM at word level to build a representation for each sentence, and another LSTM at sentence level to build representations of each whole dialogue, which is then fed to a binary classifier.
   \end{comment}
      Higher values of adversarial success
      and  {\it machine-vs-random} accuracy 
       indicate the superiority of a model. We refer readers to \newcite{add} for more details.
\begin{table}
\small
\centering
\begin{tabular}{ccc}
\hline
Model&SBS&Length prediction Q\\\hline
BLEU&1.45&1.64 \\\
{\it AdverSuc}&0.034&0.040 \\ 
{\it machine-vs-random}&0.923&0.939\\\hline
\end{tabular}
\caption{Comparison of the proposed algorithm with length prediction and the standard beam search algorithm.}
\label{length-predictor}
\end{table}
Table \ref{length-predictor} presents the quantitative results: adding 
predictor rankers increases the general quality of generated responses. 

Sampled responses (from a random batch, without cherry-picking) are shown in Table \ref{sample-length}, with more examples shown in Table \ref{length20} in the Appendix. 
We force the decoding model to generate 20 tokens using the strategy described  above. 
We can clearly identify problems with the standard beam search algorithm: the decoder produces tokens that are optimal for shorter sequences, eliminating candidates from the beam that would lead to longer possibilities. Once the length reaches the point where the intended shorter sequence would naturally conclude, it has no option but to fill space with repetitions of tokens  (e.g., {\it ``go, go, go''} or strings of punctuation) and phrases (e.g., {\it i don 't know who i am or where i am or where i am or where i am}), or addenda that are sometimes contradictory (e.g., {\it it was the only way to make it to the top of the island . . . . . . but it wasn 't}). 
This issue is alleviated by the proposed length-prediction algorithm, which plans ahead and chooses tokens that lead to meaningful sequences with the desired length. 
\begin{comment}
By comparing Appendix table \ref{length20} and \ref{length30}, we can see that there is more significant quality boost for longer sequences (sequences of length 30 against 20), which fits with our expectation: the longer sequences the model wants to generate, the more necessity to plan beforehand using the future prediction term. 
\end{comment}
More coherent responses are observed when the hyperparameter $\lambda$ is set to 1 than when it is set to 5, as expected, since the decoding algorithm deviates more from the pre-trained model when $\lambda$ takes larger values.

\section{Q for Mutual Information } \label{sec:mmi}
\subsection{Background}
Maximum mutual information (MMI) has been shown to be better than maximum likelihood estimation (MLE) as an decoding objective  for conversational response generation  tasks \cite{li2016diversity}. 
The mutual information between source $X$ and target $Y$ is given by $\log [p(X,Y)/p(X)p(Y)]$, which measures bidirectional dependency between sources and targets, as opposed to the unidirectional dependency of targets on sources in the maximum likelihood objective. 
Modeling the bidirectional dependency between sources and targets reduces the prevalence of generic responses and leads to more diverse and interesting conversations.\footnote{This is because although it is  easy to produce a sensible generic response $Y$ regardless of the input sequence $X$, it is much harder to guess $X$ given $Y$ if $Y$ is generic.}
Maximizing a weighted generalization of mutual information between the source and target can be shown using Bayes' rule to be equivalent to maximizing a  linear combination of the forward probability $\log p(Y|X)$ (the standard objective function for \sts models) and the backward probability $\log p(X|Y)$:\footnote{When using this objective, $p(Y|X)$ and $p(X|Y)$ are separately trained models with different sets of parameters. }
\begin{equation}
Y^* = \argmax_{Y} \log p(Y|X)+\lambda\log p(X|Y)
\label{mutual}
\end{equation}
Unfortunately, direct decoding using Eq.\ref{mutual} is infeasible, 
since it
requires completion of target generation before
$p(X|Y)$ can be effectively computed, and
the enormous search space for target $y$ prevents exploring
all possibilities.
An approximation
approach is commonly adopted, in which 
an N-best list is first generated based on $p(Y|X)$ and then reranked by adding $p(X|Y)$.
The problem with this reranking strategy is that
the beam search step gives higher priority to optimizing the forward probability,
resulting in solutions that are not globally optimal. 
% More seriously, sequences can only be reranked when decoding is completed. 
Since hypotheses in beam search are known to lack diversity \cite{li2016simple,vijayakumar2016diverse,gimpel2013systematic},
after decoding is finished, it is sometimes too late for the reranking model to have significant impact. 
\newcite{shao15} confirm this problem and show that the reranking approach helps for short sequences but not longer ones. 
\subsection{Training Q for Mutual Information}
The first term of  Eq.~\ref{mutual} is the same as standard \sts decoding. We thus focus our attention on the second term, $\log p(X|Y)$. 
To incorporate the  backward probability into intermediate decoding steps, we use a model to estimate the future value of $p(X|Y)$ when generating each token $y_t$. 

For example, suppose that we have a source-target pair with source $X={}${\it ``what 's your name''} and target $Y={}${\it ``my name is john''}. 
The future backward probability of the partial sequences {\it ``my''}, {\it ``my name''}, {\it ``my name is''}  
is thus $p(X|Y)$. Again, we use $Q(y_t)$ to denote the function that maps a partially generated sequence to its future backward probability, and we can  factorize Eq.~\ref{mutual} as follows:
\begin{equation}
y_t=\argmax_{y}~ \log p(y_{1:t-1}, y|X)+\lambda Q(y)
\label{eq:mmi}
\end{equation}

We propose two ways to obtain the future backward-probability estimation function $Q(y_t)$. 

(1) As in the strategies described in Sections~\ref{sec:bleu} and \ref{sec:length}, we first pretrain a \sts model for both $p(Y|X)$
and $p(X|Y)$.
 The training of the latter is the same as a standard  \sts model but with sources and targets swapped.  Then we train an additional future backward-probability estimation function $Q(X,y_{1:t})$, which takes as inputs 
the hidden representation of intermediate decoding steps (i.e., $h_t$) from the forward probability model
 and predicts the backward probability for the entire target sequence $Y$ using the pretrained backward \sts model (i.e., $\log p(X|Y)$ with $Y$ being the full target).

\begin{table}
\small
\centering
\begin{subtable}{.5\textwidth}
\centering
\begin{tabular}{cccc}
\hline
&Q for MMI&MMI&SBS\\\hline
BLEU&1.87&1.72&1.45\\
AdverSuc&0.068&0.057&0.043\\
Distinct-1&0.019&0.010&0.005\\
Distinct-2&0.058&0.030&0.014\\\hline
\end{tabular}
\caption{Full dataset.}
\label{tab:mmi:full}
\end{subtable}

\begin{subtable}{.5\textwidth}
\centering
\vspace{1em}
\begin{tabular}{cccc}
\hline
&Q for MMI&MMI&SBS\\\hline
BLEU&2.13&2.10&1.58\\
AdverSuc&0.099&0.093&0.074\\
Distinct-1&0.024&0.014&0.007\\
Distinct-2&0.065&0.033&0.017\\\hline
\end{tabular}
\caption{Set with short targets.}
\label{tab:mmi:short}
\end{subtable}
\begin{subtable}{.5\textwidth}
\centering
\vspace{1em}
\begin{tabular}{cccc}
\hline
&Q for MMI&MMI&SBS\\\hline
BLEU&1.52&1.34&1.58\\
AdverSuc&0.042&0.029&0.022\\
Distinct-1&0.017&0.008&0.004\\
Distinct-2&0.054&0.027&0.012\\\hline
\end{tabular}
\caption{Set with long targets.}
\label{tab:mmi:long}
\end{subtable}
\caption{Comparison of the proposed future prediction model with MMI-based reranking ({\it MMI}) and MLE-based standard beam search ({\it SBS}).}
\label{tab:mmi}
\end{table}
(2) We 
can directly 
train models to calculate  $Q(y_t)=p(X| y_{1:t})$, i.e., the probability of generating a full source given a partial target.
To do this, we first break $y$ into a series of partial sequences, i.e., $y_{1:1}$, $y_{1:2}$, ..., $y_{1:N}$,  
which is 
$\{${\it ``i''}, {\it ``i am''}, {\it ``i am john''}$\}$ in the example above. 
Then we pair each partial sequence $y_{1:t}$ ($1\leq t\leq N$) with the source and use each pair $(y_{1:t},X)$ as a training example to a train \sts model, with $y_{1:t}$ as the source and $X$ as the target. 
Since we are increasing the size of the training set by roughly a factor of 10 (the average target length is about 10), training is extremely computation-intensive. 
We reduce the training time by grouping $y_{1:t}$ by length and training a separate \sts model for each length group. 
At decoding time, we use the score from the model corresponding to the length of the current partially decoded target to generate the next token. Since \sts models for different target lengths are independent, they can be trained in parallel. 

We find that option (2) generally outperforms option (1), but option (2)  requires training in parallel on a large number of machines.

\begin{table}
\small
\centering
\begin{tabular}{p{2cm}p{5cm}}\hline
{\bf Input}&it s an 8 symbol address .\\
{\bf SBS}&i don 't know where it came from .\\
{\bf MMI}&i don t know where it came from .\\
{\bf Q(MMI) ($\lambda$=1)}&i don 't know the exact address .\\
{\bf Q(MMI) ($\lambda$=5)}&it 's a very old house in paris\\\hline
{\bf Input}&that 's hardly a winning attitude .\\
{\bf SBS}&that 's the way it 's always been .\\
{\bf MMI}&it 's the only way to win the race .\\
{\bf Q(MMI) ($\lambda$=1)}&it 's just a game .\\
{\bf Q(MMI) ($\lambda$=5)}&it 's just a game of golf . take it easy .\\\hline
{\bf Input}&what is up with those two ?\\
{\bf SBS}&i don 't know what 's going on .\\
{\bf MMI}&i don 't know what the hell 's going on here !\\
{\bf Q(MMI) ($\lambda$=1)}&why don 't you go back and have a look ?\\
{\bf Q(MMI) ($\lambda$=5)}&why don 't we go up there now and check ?\\\hline
{\bf Input}&i got to get them in .\\
{\bf SBS}&i don 't have time for this .\\
{\bf MMI}&i don 't have time for them .\\
{\bf Q(MMI) ($\lambda$=1)}&you don 't have to worry about them .\\
{\bf Q(MMI) ($\lambda$=5)}&i gotta talk to them all .\\\hline
{\bf Input}&i see and you know why ?\\
{\bf SBS}&i don 't know what you talking about .\\
{\bf MMI}&because i 'm in love with you .\\
{\bf Q(MMI) ($\lambda$=1)}&because i don 't know what the hell i 'm talking about !\\
{\bf Q(MMI) ($\lambda$=5)}&because it 's because of you\\\hline
\end{tabular}
\caption{Sample of responses generated by (1) standard beam search ({\it SBS}); (2) the MMI reranking approach of \newcite{li2016diversity}, which performs reranking only after decoding is complete  (denoted by {\it MMI}); and (3) the  future prediction model Q(MMI) with different values of future prediction weight $\lambda$.  
}
\label{backward}
\end{table}

\subsection{Experimental Results}
We compare the results for the  approach  with standard beam search using the MLE objective and the MMI reranking approach of \newcite{li2016diversity},
which performs reranking only after decoding is finished. 
We report BLEU scores and {\it AdverSuc} scores\footnote{The {\it machine-vs-random} scores for the three models are very similar, respectively 0.947, 0.939, 0.935. } on the test set. We also report diversity scores (denoted by {\it Distinct-1} and {\it Distinct-2}); these are defined as in \newcite{li2016diversity} to be the 
the number of distinct unigrams and bigrams (respectively)
in generated responses, divided by the total
number of generated tokens (to avoid favoring long
sentences).
Additionally, we split the dev set into a subset containing longer targets (with length larger than 8) and a subset containing shorter ones (smaller than 8). During decoding, we force the model to generate targets of the same length as the gold standard targets using the strategy described in Section~\ref{length-e}.

Table \ref{tab:mmi} presents quantitative results for the different decoding strategies.  On the full test set, the 
future backward-probability prediction 
 model  outperforms the approach of reranking when decoding is fully finished. Specifically, a larger performance improvement is observed on examples with longer targets than on those with shorter ones. This effect is consistent with the intuition that for short responses, due to the relatively smaller search space, doing reranking at the end of decoding is sufficient, whereas this is not the case with longer sequences: as beam search proceeds, a small number of prefixes gradually start to dominate, with hypotheses differing
only in punctuation or minor morphological variations. 
Incorporating mutual information in the early stages of decoding maintains diverse hypotheses, leading to better final results. 

Table \ref{backward} presents sampled outputs from each strategy, with more results shown in the Table \ref{backward-appendix} (Appendix). 
As can be seen, the
results from
reranking are generally better than those from MLE, but sometimes both approaches still generate the same generic outputs. This is due to the fact that reranking is performed only after more interesting outputs have fallen off the beam.  
Using smaller values of $\lambda$, the  future backward-probability prediction approach generally yields better results than reranking. When using larger values of $\lambda$, the algorithm tends to produce more diverse and interesting outputs but has a greater risk of generating irrelevant responses. 

\section{Q for BLEU/ROUGE} \label{sec:bleu}
The future outcome function $Q$ can be trained to predict arbitrary features. These features include 
 BLEU \cite{papineni2002bleu} or ROUGE \cite{lin2004rouge} scores. 
We thus train $Q$ to directly predict future BLEU or ROUGE values. 
In this situation, 
the future prediction function is able to reduce the discrepancy between training (using maximum likelihood objective) and testing (using BLEU or ROUGE) \cite{wiseman2016sequence,shen2015minimum,marc2016sequence}. 
\subsection{Model}
Given a pre-trained sequence generation model,  an input sequence $X$, and a partially decoded sequence $y_{1:t-1}$, we want to estimate the future reward for taking the action of choosing word $y_t$ for the current time-step. We denote
this estimate $Q(\{y_t, y_{1:t-1},X\})$, abbreviated $Q(y_t)$ where possible.

The future prediction network is trained as follows:
we first sample $y_t$ from the distribution $p(y_t|X,y_{1:t-1})$, then decode the remainder of the sequence $Y$ using beam search. 
The future outcome for the action $y_t$
is thus the score of the final decoded sequence, $q(Y)$. 
Having obtained pairs $(q(Y),\{X, y_{1:t}\})$, we train a neural network model that takes as input $X$ and $y_{1:t}$ to predict $q(Y)$. 
The network first maps the input sequence $X$ and the partially decoded  sequence $y_{1:t}$ to vector representations using LSTMs, and then uses another network that takes the concatenation of the two vectors to output the final outcome $q(Y)$.
The future prediction network is optimized by 
minimizing the mean squared loss between the predicted value and the real $q(Y)$ during training. 

At decoding time, 
$Q(y_t)$ is  incorporated into the decoding model to push the model to take actions that lead to better future outcomes. 
An action $y_t$ is thus evaluated by the following function:
\begin{equation}
y_t=\argmax_{y}~ \log p(y_{1:t-1}, y|X)+\lambda Q(y)
\label{eq:fut_pred}
\end{equation}
$\lambda$ is a hyperparameter that is tuned on the development set. 

\subsection{Experiments}
We evaluate the decoding model on two sequence generation tasks, machine translation and abstractive summarization.

\paragraph{Machine Translation} We use  the  German-English
machine translation track of the IWSLT 2014 \cite{cettolo2014report}, which 
consists of sentence-aligned subtitles of TED and TEDx talks. For fair comparison, we followed exactly the data processing protocols defined in \newcite{marc2016sequence}, which have also been adopted by \newcite{bahdanau2016actor} and \newcite{wiseman2016sequence}. 
The training data consists of roughly 150K sentence pairs,
in which the average English sentence is 17.5 words long and the average German sentence is 18.5
words long.
The test set
is a concatenation of dev2010, dev2012, tst2010, tst2011 and tst2012, consisting of 6750 sentence
pairs.
 The English dictionary has 22822 words, while the German has 32009 words.

We train two models, a vanilla LSTM \cite{sutskever2014sequence} and an attention-based model \cite{bahdanau2014neural}.
For the attention model, we use the \emph{input-feeding} model described in 
\newcite{luong2015effective}  with one minor modification:  
the weighted attention vectors that are used in the softmax token predictions and those fed to the recurrent net at the next step use different sets of parameters. Their values can therefore be different, unlike in \newcite{luong2015effective}. We find that this small modification significantly improves the capacity of attention models, yielding more than a +1.0 BLEU score improvement. 
We use structure similar to that of \newcite{wiseman2016sequence}, a single-layer sequence-to-sequence model with 256 units for each layer. 
We use beam size 7 for both  standard beam search (SBS) and future outcome prediction.
\begin{table}
\centering
\small
\begin{tabular}{ll}\hline
REINFORCE (Ranzato et al., 2016)&20.7 \\
Actor-Critic \cite{bahdanau2016actor}  &22.4 \\
\newcite{wiseman2016sequence}&26.3 \\\hline
vanilla LSTM + SBS& 18.9 \\
vanilla LSTM + Q(BLEU) &19.7 (+0.8)\\\hline
attention+ SBS&27.9 \\
attention + Q(BLEU) &{\bf 28.3} (+0.4) \\\hline
\end{tabular}
\caption{BLEU scores for different systems. Baseline scores are best scores reprinted from corresponding papers. SBS denotes standard beam search.}
\label{MT}
\end{table}

\begin{table}
\centering
\small
\begin{tabular}{ll}\hline
attention + SBS&12.2\\
attention + Q(ROUGE) &{\bf 13.2} (+1.0) \\\hline
\end{tabular}
\caption{ROUGE-2 for abstractive summarization. SBS denotes standard beam search.}
\label{ROUGE}
\end{table}

Results are shown in Table \ref{MT}, with {\it SBS} standing for the
standard beam search model and {\it future func} as the proposed future prediction model.
Baselines employed include the REINFORCE model described in \newcite{marc2016sequence}, the actor-critic RL model described in \newcite{bahdanau2016actor}
and  the beam-search
training scheme described in \newcite{wiseman2016sequence}. Results are reprinted from the best setting in the corresponding paper.

Our implementation of the attention model itself already achieves state-of-the-art performance on this benchmark. The proposed future outcome model adds +0.4 BLEU, pushing the SOTA performance up to 28.3. 
Since the trained \sts model is already quite strong, there is less room for improvement.
For the vanilla LSTM, however, due to its relative inferiority, we observe a more significant improvement from the future outcome prediction approach. 
\paragraph{Abstractive Summarization}
We follow the protocols described in \newcite{rush2015neural}, in which  the source input is the first sentence of a new article and 
the target output is the headline. Our training dataset consists of 2M pairs. 
We train a two-layer word-level  attention  model with 512 units for each layer. 
Experimental results are shown in Table \ref{ROUGE}. We observe a +1.0 ROUGE performance improvement from the proposed model over standard beam search.

\section{Conclusion} \label{sec:conclusion}
In this paper, we propose a
general strategy that enables a neural decoder to generate outputs that have specific properties of  interest. 
We show how to use a  model Q to optimize three useful properties of the output---sequence length,  mutual information and BLEU/ROUGE scores 
---and investigate the effects of different designs for the
predictor model and decoding algorithm. 
Our model provides a general and easy-to-implement way  to control neural generation models to meet their specific needs,
while improving results on a variety of generation tasks.

\bibliographystyle{acl_natbib}

\bibliography{MMI_MT}
{\bf Appendix}

\begin{table*}
\small
\centering
\begin{tabular}{ll}\hline
{\bf Input}&one of those american robot cars .\\
{\bf SBS}&i don 't know what the hell you 're talking about ... ... but it 's the only solution .\\
{\bf FP($\lambda$=1)}&you know i don t know if you 've got a lot of good ideas on the subject sir !\\
{\bf FP($\lambda$=5)}&the first time they 've ever had a chance to know ... ... who got them killed ? ! !\\\hline
{\bf Input}&the comanches mate their women early .\\
{\bf SBS}&they don 't even know who they 're dealing with but they don 't want to see their children .\\
{\bf FP($\lambda$=1)}&and now it seems to have a new owner who has a name for a new friend in america .\\
{\bf FP($\lambda$=5)}&the only reason that you don 't have a problem is that there 's no more . . . .\\\hline
{\bf Input}&but she is still our innocent .\\
{\bf SBS}&she 's not the only one who knows what she 's doing to her own life . . . .\\
{\bf FP($\lambda$=1)}&and now we can all find out what happened in the last place that she was in love with isn\\
{\bf FP($\lambda$=5)}&and the only reason we 've taken the case ... is because she has no intention of leaving her .\\\hline
{\bf Input}&oh must i the whole sentence ?\\
{\bf SBS}&i don 't know if i 'm guilty or not guilty ... ... but i 'm guilty of murder .\\
{\bf FP($\lambda$=1)}&i mean i 'll have you in my quarters when i 'm through with the crime scene . ! ?\\
{\bf FP($\lambda$=5)}&well if you 'd mind saying i 'd rather be the one who 's on my own now ! !\\\hline
{\bf Input}&because that 's a real guy .\\
{\bf SBS}&that 's why he 's in love with a guy who doesn 't know what he 's talking about .\\
{\bf FP($\lambda$=1)}&but i think we all know who he was and why we came from the real world ! ? !\\
{\bf FP($\lambda$=5)}&i mean you know who i think that 's the guy who lives in the real world right now .\\\hline
{\bf Input}&that 's what this job is .\\
{\bf SBS}&you don 't have to worry about the money ... ... or the money or the money or anything .\\
{\bf FP($\lambda$=1)}&and i 'll tell you that i 'm not gonna let you in on this one okay ? ! ?\\
{\bf FP($\lambda$=5)}&it means we have to go to jail because we have to go through the whole thing all right sir\\\hline
{\bf Input}&supervisor tang they are starting to quarrel\\
{\bf SBS}&i don 't want to be late for the meeting . . . . . . but i can 't\\
{\bf FP($\lambda$=1)}&but the boss doesn 't even care about his family and he 's the only one in this family .\\
{\bf FP($\lambda$=5)}&and if we do not succeed ... ... the next time we 'll be back together . . . !\\\hline
{\bf Input}&we 're almost out of time .\\
{\bf SBS}&we 've got to get back to the ship before the sun hits the moon and the moon will rise .\\
{\bf FP($\lambda$=1)}&and the next day the next time you go into the city you 'll go home to bed again !\\
{\bf FP($\lambda$=5)}&the last time i checked out ... ... i had to be in the wrong place . ! ? !\\\hline
{\bf Input}&get your fat ass over there !\\
{\bf SBS}&get the fuck out of my way ! ! ? ! ? ! ? ! ? ! ? ! ?\\
{\bf FP($\lambda$=1)}&don 't you think i 've got a better idea than to get my ass kicked out here ? !\\
{\bf FP($\lambda$=5)}&you know i 'm going to have to go over and see the other two kids ? ! ? !\\\hline
{\bf Input}&i gotta have the whole beehive .\\
{\bf SBS}&i 'm in the middle of the street and i 'm in the middle of the street and i say\\
{\bf FP($\lambda$=1)}&he says i gotta go to bed and i 'll get the car and i 'll make it easy !\\
{\bf FP($\lambda$=5)}&the other night when the party starts going down ... ... there 's no other place for them ! ?\\\hline
{\bf Input}&what are you trying to hide ?\\
{\bf SBS}&i don 't know what you 're talking about . . . . . . but it 's the truth .\\
{\bf FP($\lambda$=1)}&i thought he said that i 'd find a way to make sure he didn 't kill me . !\\
{\bf FP($\lambda$=5)}&i thought he said you were gonna go to a meeting ... ... or a party ? ! ? !\\\hline
{\bf Input}&oh my that looks yummy lily .\\
{\bf SBS}&i don 't think she 's gonna wear it in the first place but she 's gonna be okay .\\
{\bf FP($\lambda$=1)}&you know i don 't think you 've ever been in love with a girl like me ever before !\\
{\bf FP($\lambda$=5)}&you know i don 't know how much longer that 's gonna be in your body . okay ? .\\\hline
{\bf Input}&the man said nothin at first .\\
{\bf SBS}&he said he didn 't know what he was talkin about but he didn 't say nothin about it .\\
{\bf FP($\lambda$=1)}&he said that the first thing that happened to him was the first time he ever came here again !\\
{\bf FP($\lambda$=5)}&and then the second and third ... ... i went to the bathroom and there 's nothin wrong ? !\\\hline
{\bf Input}&we get in a big hand .\\
{\bf SBS}&we don 't know what we 're talking about but we 're in the middle of a big fight .\\
{\bf FP($\lambda$=1)}&the other day when we came here ... ... we had to go to the police right away . !\\
{\bf FP($\lambda$=5)}&you know we 'll see you around the block when we get there and get your ass outta here !\\\hline
{\bf Input}&i have an announcement to make .\\
{\bf SBS}&i want you to meet the president of the united states ... ... and i want to thank you .\\
{\bf FP($\lambda$=1)}&you know that i was the first person i ever worked with in my entire career . . . !\\
{\bf FP($\lambda$=5)}&you know that i was the only man in my crew who would be interested in this . . .\\\hline
{\bf Input}&the unknown order is chaos man .\\
{\bf SBS}&it 's the only thing we need to do right now right now is that we 're in the middle .\\
{\bf FP($\lambda$=1)}&you know what we 'll find on the planet when you 're in the middle of the war ? !\\
{\bf FP($\lambda$=5)}&i know the right way to get rid of it but i can 't handle it . ! ? !\\\hline
\end{tabular}

\caption{Sampled examples from responses generated by standard beam search (denoted by {\it SBS}) and the proposed Length-Predictor algorithm in Section 4 with different weight $\lambda$. We force each decoding algorithm to generated responses with length 20. We randomly take one batch from the dev set and the examples are not cherry picked.}
\label{length20}
\end{table*}

\begin{table*}[!htb]
\centering
\footnotesize
\begin{tabular}{p{2cm}p{5cm}p{2cm}p{5cm}}\hline
{\bf Input}&we 're not at war sir .\\
{\bf MLE}&we 're in the middle of a war .&{\bf MMI}&we 're in the middle of a war .\\
{\bf FP ($\lambda$=1)}&we don 't know where we are .&{\bf FP ($\lambda$=5)}&we 're not at peace captain we 'll do what 's best for them\\\hline
{\bf Input}&the agency doesn 't think so .\\
{\bf MLE}&you don 't know what you 're talking about .&{\bf MMI}& oh , really ?\\
{\bf FP ($\lambda$=1)}&i don 't think it 's funny .&{\bf FP ($\lambda$=5)}&i think it 's because they don 't know who he is .\\\hline
{\bf Input}&he just stole her from me .\\
{\bf MLE}&i don 't know who he is .&{\bf MMI}&i don 't know who he is .\\
{\bf FP ($\lambda$=1)}&calm down ! !&{\bf FP ($\lambda$=5)}&oh, what 's your plan then ?\\\hline
{\bf Input}&the comanches mate their women early .\\
{\bf MLE}&that 's the way it 's gonna be .&{\bf MMI}&i don 't know what the hell they 're talking about but they don 't care about women .\\
{\bf FP ($\lambda$=1)}&that 's why they come here .&{\bf FP ($\lambda$=5)}&they 're a bunch of wild savages ...\\\hline
{\bf Input}&there 's no tape inside anyway !\\
{\bf MLE}&no no no no no no no no no no no no no no no no no no .&{\bf MMI}&no no no no no no no no no no no no no no no no no no !\\
{\bf FP ($\lambda$=1)}&where did he hide the tape ?&{\bf FP ($\lambda$=5)}&i don 't think i should take you to the movies !\\\hline
{\bf Input}&supervisor tang they are starting to quarrel\\
{\bf MLE}&why don 't you come with me ?&{\bf MMI}&why don 't you go back to your room and have a chat with them\\
{\bf FP ($\lambda$=1)}&this is ridiculous !&{\bf FP ($\lambda$=5)}&i don 't care !\\\hline
{\bf Input}&we 're almost out of time .\\
{\bf MLE}&there 's no need to worry about it .&{\bf MMI}&
i 'll be right back\\
{\bf FP ($\lambda$=1)}&there 's no need to panic .&{\bf FP ($\lambda$=5)}& we 've got to get back to the ship before it 's too late ..\\\hline
{\bf Input}&get your fat ass over there !\\
{\bf MLE}&get the fuck out of here !&{\bf MMI}&get the fuck out of my face !\\
{\bf FP ($\lambda$=1)}&get the fuck out of my face !&{\bf FP ($\lambda$=5)}&don 't you get tired of this man 's fucking shit !\\\hline
{\bf Input}&oh my that looks yummy lily .\\
{\bf MLE}&i don 't think it 's funny .&{\bf MMI}&i don 't think it 's a good idea to wear a wedding dress .\\
{\bf FP ($\lambda$=1)}&that 's a nice dress .&{\bf FP ($\lambda$=5)}&i thought we agreed to have dinner together with your wife and your daughter and ...\\\hline
{\bf Input}&our scout ships have reached UNknown\\
{\bf MLE}&you know what you doing, right ?
&{\bf MMI}&there 's no sign of the rebel fleet ..\\
{\bf FP ($\lambda$=1)}&the rebel ships are still in orbit .&{\bf FP ($\lambda$=5)}&they 're still in orbit in the alpha quarter of space ...\\\hline
{\bf Input}&the UNknown order is chaos man .\\
{\bf MLE}&it 's the only way to get rid of it .&{\bf MMI}&it 's the only way to get rid of it .\\
{\bf FP ($\lambda$=1)}&it 's just the beginning .&{\bf FP ($\lambda$=5)}&it 's the rules of the country\\\hline
{\bf Input}&but you are not a mistake .\\
{\bf MLE}&i don 't know who you are   or what you are.&{\bf MMI}&i don 't know who you are or what you are or what you are .\\
{\bf FP ($\lambda$=1)}&i don 't know who you are .&{\bf FP ($\lambda$=5)}&you have no right to give your life for the cause of the mistake of our enemy ...\\\hline
{\bf Input}&heaven and earth must i remember ?\\
{\bf MLE}&i don't know what you are talking about&{\bf MMI}&there 's no such thing as ghosts .\\
{\bf FP ($\lambda$=1)}&there 's no need to be afraid of the dark .&{\bf FP ($\lambda$=5)}&there 's only one place where we can see it\\\hline
{\bf Input}&we got a murder down here ...&
{\bf SBS}&i don 't know what the hell 's goin on here .\\
{\bf MMI}&i don 't know what the hell 's goin on here .&
{\bf FP ($\lambda$=1)}&i don 't know who he is ...\\
{\bf FP ($\lambda$=5)}&we gotta do something about this ...\\\hline
{\bf Input}&but venus was no easy target .\\
{\bf SBS}&there 's no need to worry about it .&
{\bf MMI}&it was the only way to survive .\\
{\bf FP ($\lambda$=1)}&i don 't know what the hell you 're talking about ...&
{\bf FP ($\lambda$=5)}&there 's a whole bunch of them in the middle east\\\hline

\end{tabular}
\caption{Sampled examples from responses generated by (1) standard beam search using MLE objective  (denoted by {\it MLE}); (2) the MMI reranking model  (denoted by {\it MMI})
in \cite{li2016diversity}
that perform reranking only after decoding is fully done; (3) the proposed partial-seq-MMI model (denoted by {\it FP}) with different values of future prediction weight $\lambda$.  
}
\label{backward-appendix}
\end{table*}

\end{document}